\documentclass[twocolumn]{article}

\usepackage{arxiv}

\usepackage[utf8]{inputenc} 
\usepackage[T1]{fontenc}    
\usepackage{hyperref}       
\usepackage{url}            
\usepackage{booktabs}       
\usepackage{amsfonts}       
\usepackage{amsmath}
\usepackage{nicefrac}       
\usepackage{microtype}      
\usepackage{cleveref}       
\usepackage{lipsum}         
\usepackage{graphicx}
\usepackage{natbib}
\usepackage{doi}
\usepackage{capt-of}

\usepackage{amssymb}
\usepackage{algorithm}
\usepackage{algpseudocode}

\title{ViTmiX: Vision Transformer Explainability\\ Augmented by Mixed Visualization Methods}

\date{}

\newif\ifuniqueAffiliation
\uniqueAffiliationtrue

\ifuniqueAffiliation 
\author{ 
    \href{https://orcid.org/0000-0002-5814-992X}{\includegraphics[scale=0.06]{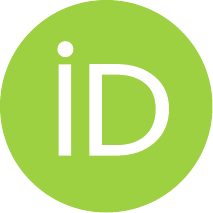}\hspace{1mm}Eduard Hogea}\thanks{eduard.hogea00@e-uvt.ro} \\
    West University of Timisoara \\
    Timisoara, Romania \\
    \texttt{eduard.hogea00@e-uvt.ro} \\
    \And
    \href{https://orcid.org/0000-0003-4846-3752}{\includegraphics[scale=0.06]{orcid.pdf}\hspace{1mm}Darian M.~Onchis} \\
    West University of Timisoara \\
    Timisoara, Romania \\
    \texttt{darian.onchis@e-uvt.ro} \\
    \And
    Ana Coporan \\
    West University of Timisoara \\
    Timisoara, Romania \\
    \texttt{ana.coporan01@e-uvt.ro} \\
    \And
    \href{https://orcid.org/0000-0001-7249-1871}{\includegraphics[scale=0.06]{orcid.pdf}\hspace{1mm}Adina Magda Florea} \\
    Politehnica University of Bucharest \\
    Bucharest, Romania \\
    \texttt{adina.florea@upb.ro} \\
    \And
    Codruta Istin \\
    Politehnica University of Timisoara \\
    Timisoara, Romania \\
    \texttt{codruta.istin@upt.ro} \\
}

\else
\fi

\renewcommand{\shorttitle}{\textit{arXiv} Template}

\hypersetup{
pdftitle={ViTmiX: Vision Transformer Explainability\\ Augmented by Mixed Visualization Methods},
pdfsubject={q-bio.NC, q-bio.QM},
pdfauthor={},
pdfkeywords={First keyword, Second keyword, More},
}

\begin{document}
\twocolumn[{
  \begin{@twocolumnfalse}

	\maketitle

	\begin{center}
		\includegraphics[width=0.5\textwidth]{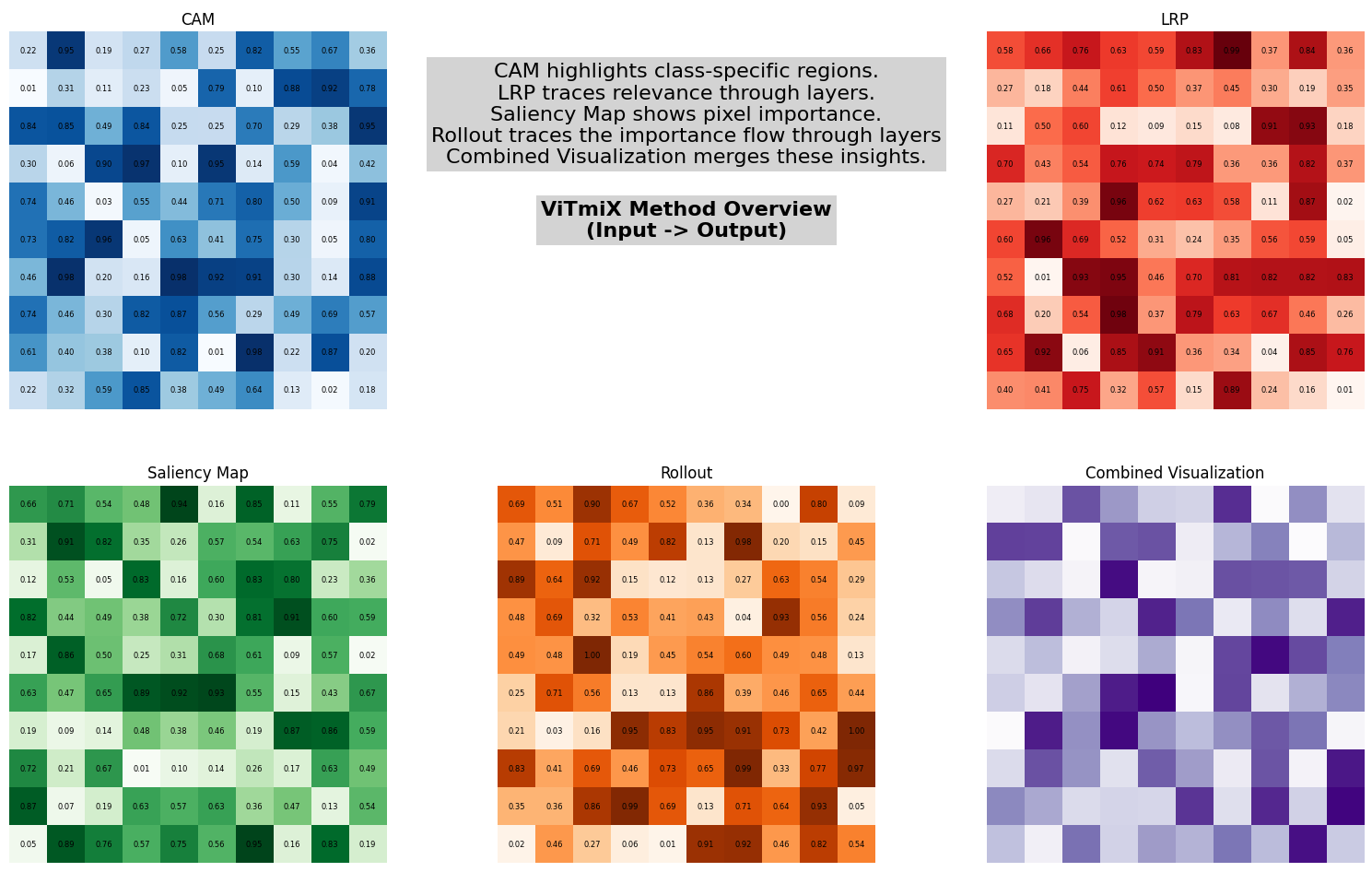}
		\captionof{figure}{\textbf{ViTmiX: Complementarity of explainable AI methods}. Each feature attribution map represents the relevance or importance scores calculated by the respective methods applied to a visual transformer model. Each cell in the heatmap visualization corresponds to the relevance or importance of a particular feature (e.g., a pixel in an image). The colors (Blues, Reds, Greens, Oranges, Purples) are used to visually differentiate between the methods (CAM, LRP, Rollout Attention, Saliency Map) and their mixed features attribution map.}
	    \label{fig:vitmix}
	\end{center}

	\vspace{1em}

	\begin{abstract}
		Recent advancements in Vision Transformers (ViT) have demonstrated exceptional results in various visual recognition tasks, owing to their ability to capture long-range dependencies in images through self-attention mechanisms. However, the complex nature of ViT models requires robust explainability methods to unveil their decision-making processes. Explainable Artificial Intelligence (XAI) plays a crucial role in improving model transparency and trustworthiness by providing insights into model predictions. Current approaches to ViT explainability, based on visualization techniques such as Layer-wise Relevance Propagation (LRP) and gradient-based methods, have shown promising but sometimes limited results. In this study, we explore a hybrid approach that mixes multiple explainability techniques to overcome these limitations and enhance the interpretability of ViT models. 
        Our experiments reveal that this hybrid approach significantly improves the interpretability of ViT models compared to individual methods. We also introduce modifications to existing techniques, such as using geometric mean for mixing, which demonstrates notable results in object segmentation tasks. To quantify the explainability gain, we introduced a novel post-hoc explainability measure by applying the Pigeonhole principle. These findings underscore the importance of refining and optimizing explainability methods for ViT models, paving the way to reliable XAI-based segmentations.
	\end{abstract}

  \end{@twocolumnfalse}
  \vspace{1em}
}]

\section{Introduction}
\label{sec:intro}
Artificial intelligence (AI) applications span various types of input, including text \cite{deng2018deep}, images \cite{pop00001}, audio \cite{pop00002}, and more. Each type of input requires specialized techniques to process and analyze effectively, which can be further integrated into a multi-modal approach. For computer vision (CV) tasks such as image classification or facial recognition, convolutional neural networks (CNNs) have been the go-to option because of their highly accurate results \cite{fuad2020features}. 

Despite the dominance of CNNs in image classification, newer architectures such as vision transformers (ViTs) are emerging and consolidating their position in the applied field. Originally developed for natural language processing (NLP) tasks because of their impressive performance, they have also been extended to the vision domain. ViT have demonstrated performance that is comparable or even superior to those of CNN architectures.\cite{dosovitskiy2020image}

To expand the knowledge behind ViT it is crucial to understand their inner working procedure and examine their explainability\cite{kashefi2023explainability}. Explainable artificial intelligence (XAI) is a domain that challenges the transparency of AI models. XAI has the promise of making AI systems more trustworthy and accountable. Not only that but it helps in better understanding the models, uncover limitations, and improve the overall performance. This endeavor ensures that AI systems are not only powerful, but also transparent, reliable and aligned with human values and expectations. With respect to ViT, XAI can help uncover which features are most important and which image regions have impacted the most in the prediction, helping to better clarify the workings of the attention mechanism. This understanding is crucial in applications like medical diagnostics, where knowing which specific image features led to a diagnosis can enhance interpretability and trust in AI-driven medical decisions.



To understand the inner work of a ViT one can choose one of the several existing approaches: visualization techniques, attention-based methods, pruning-based methods, inherently explainable models, etc. A comprehensive review of ViT explainability is given by \cite{kashefi2023explainability}. They do a thorough organization of the existing methods, taking into consideration factors like motivation, structure, and application scenarios.

Despite the promise of ViTs, our research also emphasizes how explainability poses significant challenges. Unlike CNNs, where convolutional layers explicitly capture hierarchical features, ViTs rely on self-attention mechanisms to integrate global context and local dependencies across image patches. This unique architecture necessitates novel approaches for interpreting how these models arrive at their predictions, highlighting which image regions and features contribute most significantly to classification outcomes. Existing research has explored various visualization techniques for ViT explainability, but individually they may lack robustness or fail to provide precise interpretations. Thus, the central problem addressed in this research is to enhance the interpretability of ViTs by exploring and mixing multiple explainability techniques. Using hybrid methods that integrate the strengths of different approaches, this papers aims to uncover meaningful insights into ViT decision-making processes, with insightful visualizations, ultimately advancing the field of XAI and paving the way for more transparent and reliable AI systems.

The primary objective of our research is to explore whether mixing different explainability methods of ViT can enhance the interpretability and provide more insightful visual explanations. A graphical illustration of the ViTmiX method in a heatmap visualization style is given in Figure \ref{fig:vitmix}, where we highlight the maximum values for the three explainable visualization methods, emphasizing their complementary advantages.
The goal is to create a more informative visualization. Different attribution methods have their strengths and weaknesses. Mixing them can make the final visualization more robust, reducing the likelihood of misleading interpretations that might arise from relying on a single method. 
To achieve this, we conducted a theoretical analysis of the prominent visualization methods: attention rollout, LRP based method and gradient based methods like CAM and SAW, analyzing their strengths and weak points. The second step was that of trial and error, mixing and matching the techniques together and trying different implementations to see if our assumption of improving interpretability was right or not.

The structure of this article begins with an overview of related works, discussing existing visualization techniques, attention mechanisms, and methods to visualize pixel contributions and image features crucial for classification.
Following this, we detail our approach, starting with a theoretical analysis of established techniques in which we highlight their strengths and weaknesses and explore novel combinations and computational strategies to enhance explainability.
In the experiments section \ref{sec:Experiments}, we validate our hypothesis by demonstrating improved interpretability through hybrid methods that mix LRP and CAM with attention rollout and saliency maps. We illustrate how these approaches uncover significant image features and enhance model transparency. Concluding our findings, we underline the strenghts of our method in optimizing ViT explainability. Through this structured approach, our study contributes to advancing the understanding and application of explainable AI in ViT models.

\section{Related Work}

 \textbf{Vision Transformers} use a self-attention mechanism as a core part of their architecture \cite{robusttransformer}. This mechanism allows the model to weigh the importance of different parts of the input image when making predictions. The self-attention mechanism computes attention scores, which indicate how much focus each part of the image receives. These attention scores provide insight into how the model processes information from different parts of the image, which is why attention is widely used for explainability. There are two methods which have been proposed for mixing the attention scores across layers, attention rollout and attention flow. Attention rollout is a technique used to aggregate attention maps across multiple layers of a transformer. It provides a cumulative view of how information is propagated and aggregated throughout the layers. Although intuitive and easy applicable, it does not give a class specific explanation.  Attention flow also aims to propagate attention scores through the layers of the transformer, and the idea behind this mechanism is to compute the maximum flow problem through a graph. The latter method is very slow, which usually leads to it being left out from comparisons with other methods.

The main visual methods for ViT explainability that have been explored in our research can be broken down into 2 different classes: (1) gradient-based and (2) attribution propagation methods. 

\textbf{Gradient-based} methods for interpreting deep learning models work by calculating the gradients of the model's output with respect to the input features. The logic is based on the idea that these gradients indicate how much each input feature contributes to the model's prediction. By visualizing these gradients, these methods highlight important regions or aspects of the input that most influence the model's decision. The result of such method is usually called a saliency map. 
Saliency maps highlight the regions in the input image that have the most significant impact on the model's prediction. Vanilla Saliency computes the absolute value of the gradients of the output with respect to the input. 

Class activation map (CAM)\cite{zhou2016learning} proposes to compute a linear combination between the weights and feature maps from the last layer of the model. Attention guided CAM is selectively aggregating gradients propagated from the classification output to each self-attention layer\cite{leem2024attention}. The method gathers contributions of image features from various locations in the input image. Additionally, these gradients are refined using normalized self-attention scores, which represent pairwise patch correlations. These scores enhance the gradients with patch-level context information identified by the self-attention mechanism. SAW\cite{mallick2022saw} is a method that is derived from the attention rollout mechanism. Instead of just combining all the attention maps together, which may result in not such specific results, they are weighting the self attention maps and are aggregating them based on their weights. The attention is being weighted at a specific class, which makes their method to be class specific.

\textbf{Attribution propagation} methods are based on the Deep Taylor Decomposition(DTD). DTD propagates relevance scores backward from the output layer to the input layer, and the one such method is called the Layerwise Relevance Propagation (LRP), on which most of the research for ViT explainability is  based. LRP can be adapted to propagate relevance through the self-attention mechanisms. Gradient based methods explain the contribution of the image features elicited through multiple layers, while LRP based methods capture the contribution of the independent pixels to the classification output\cite{leem2024attention}. A class specific LRP based visualization technique has been proposed that uses LRP based relevance to calculate scores for each attention head in every layer of the model\cite{chefer2021transformer}. These scores are then integrated throughout the attention graph by combining relevance and gradient information, iteratively eliminating negative contributions.

The possibility of combining different types of XAI methods was proposed as a manifesto idea in the paper \cite{manifesto}. A hybrid saliency enhancing method, with scaling and sliding, was explored in \cite{saliency}. It improves saliency by fusing saliency maps extracted from multiple patches at different scales from different areas, and combines these individual maps using a novel fusion scheme that incorporates channel-wise weights and spatial weighted average. Another idea of mixing XAI models was explored in the paper \cite{mixing}. The authors perform a detailed evaluation of XAI through a mixed methods approach, combining quantitative standardised scales and qualitative techniques.

\section{Method}
\label{sec:approach}

Understanding the strengths and weaknesses of interpretability methods in deep learning is crucial for effectively analyzing model decisions. Attention rollout techniques help understand how attention evolves across layers, providing dynamic insights into the overall decision-making process of ViT. While not inherently class-specific and prone to noise, mixing rollout with other methods can enhance localization accuracy. Saliency maps are straightforward and efficient to compute by taking gradients of the output with respect to the input, facilitating efficient pixel-wise visualizations of input features. Yet, because of their simplicity they may suffer from diffuse outputs, lacking clear indications of exact features. Class Activation Mapping  methods are easy to implement across various architectures, but their weakness is that the gradients that they use for localization might capture both relevant and irrelevant features, making it challenging to discern which regions are genuinely important for classification. Because they usually rely on the final convolutional layer activations, they might not capture the information across all the layers, limiting the quality and accuracy of the generated heatmaps. LRP methods provide a concrete understanding of model decisions. They are taking into consideration all the layers of the model by propagating relevance scores across the layers of the transformer. Just because of their precise explanation, by highlighting only specific areas of the target class, you can't always see the bigger picture.

In our visual classification experiments, we utilize a pretrained ViT-based model\cite{dosovitskiy2020image}. The model adopts a BERT-like architecture. The images used are of size 224x224, the size of the patches are of 16, and they are followed by flattening and linear transformations to generate a sequence of vectors. A classification token is prepended to the sequence and serves as the input for classification tasks.

\textbf{Dataset:} We used the Pascal Visual Object Classes (VOC) dataset to test the potential of our visualization methods for segmentation tasks \cite{voc}. The Pascal VOC 2012 dataset is a well known benchmark in visual object category recognition and detection, containing annotated images for tasks such as classification, detection, segmentation, and action recognition. 


\subsection{Evaluation Metrics}
In addition to directly evaluating the generated heatmaps, we employed specific metrics aimed at assessing the performance of the explainability methods. Key among these metrics are segmentation metrics such as Pixel Accuracy, F1 Score, and Jaccard Index. These metrics are essential for evaluating the quality of segmentation results, providing insights into how accurately and precisely the models can identify and localize relevant features in input images. The heatmaps produced were processed using OTSU thresholding to create binary masks. These resulting masks were then compared with ground truth masks to evaluate performance. Pixel Accuracy quantifies the overall correctness of the segmentation predictions. The F1 Score offers a balanced measure of precision and recall across different classes, highlighting the trade-off between these two aspects. The Jaccard Index (or Intersection over Union) measures the overlap between predicted segments and the ground truth, offering a comprehensive view of segmentation performance.


\begin{figure}[!t]
\centering
\includegraphics[width=0.5\textwidth]{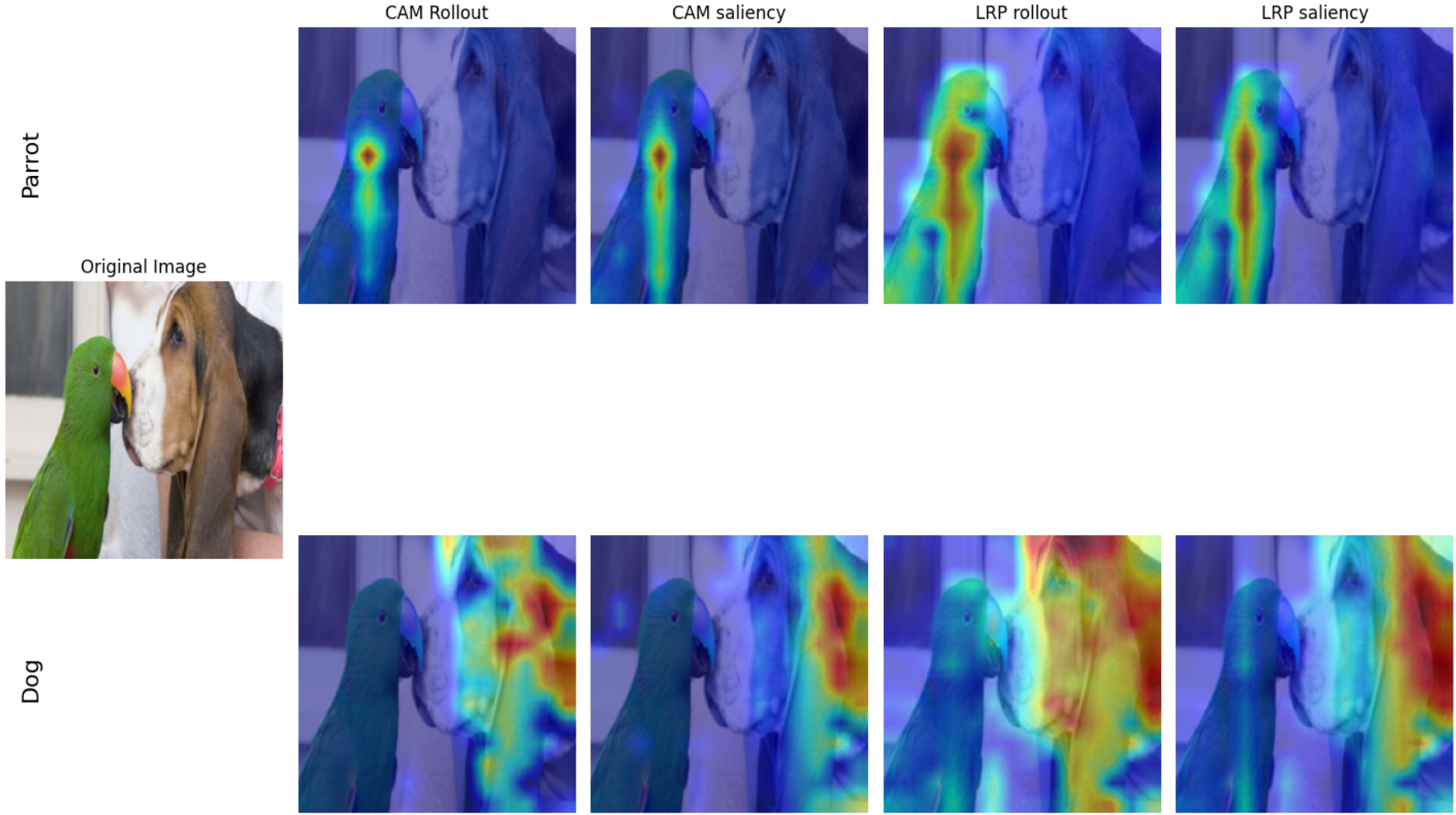}
\caption{Heatmaps of proposed visualization techniques — Multiplication}
\label{fig:plot3}
\end{figure}


\begin{figure}[t]
\centering
\includegraphics[width=0.3\textwidth]{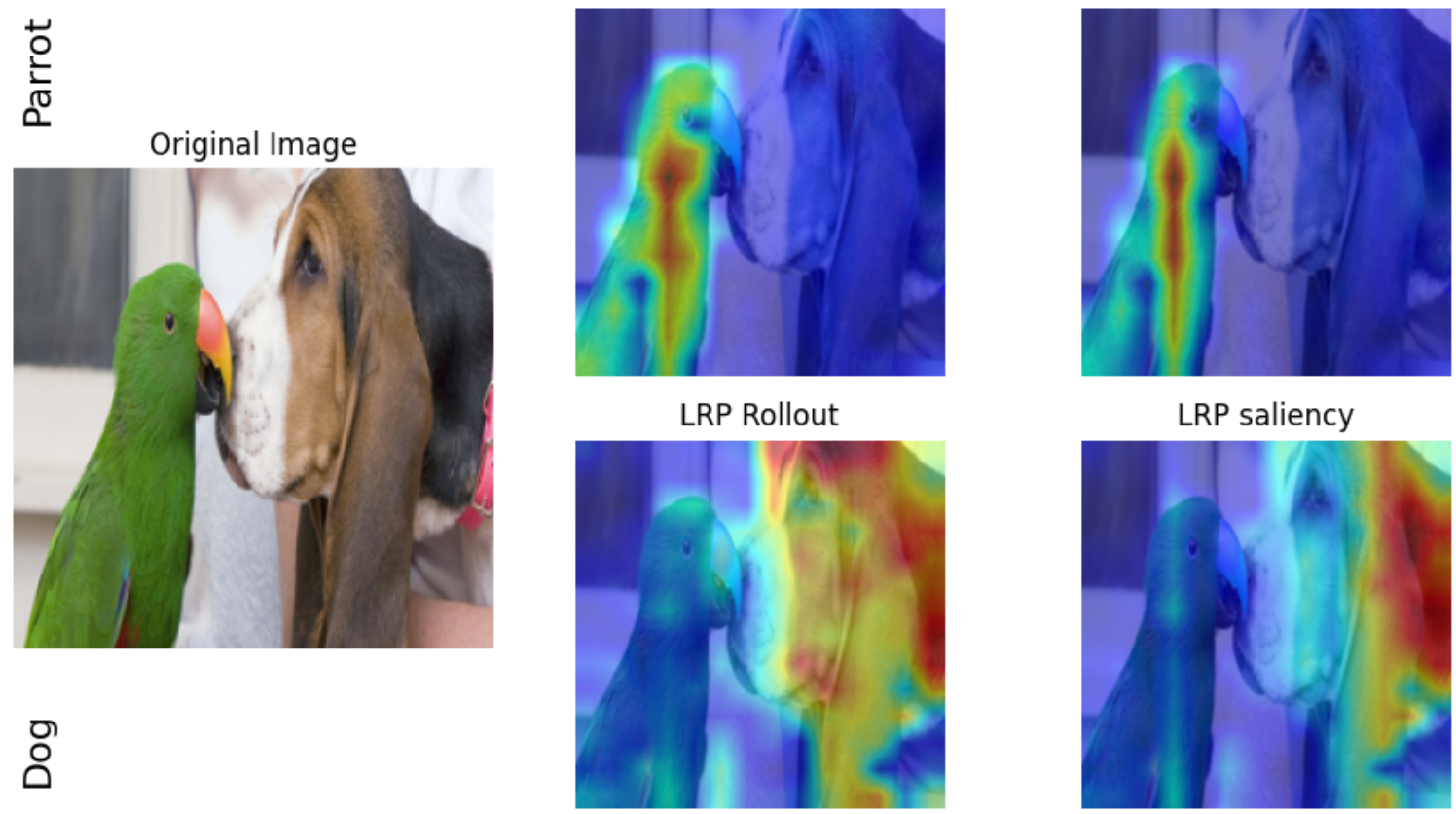}
\caption{Heatmaps of proposed visualization techniques — Geometric Mean}
\label{fig:plot4}
\end{figure}

\subsection{Experimental assessment}
\label{sec:Experiments}


\begin{figure}[!t]

\includegraphics[width = 0.4\textwidth]{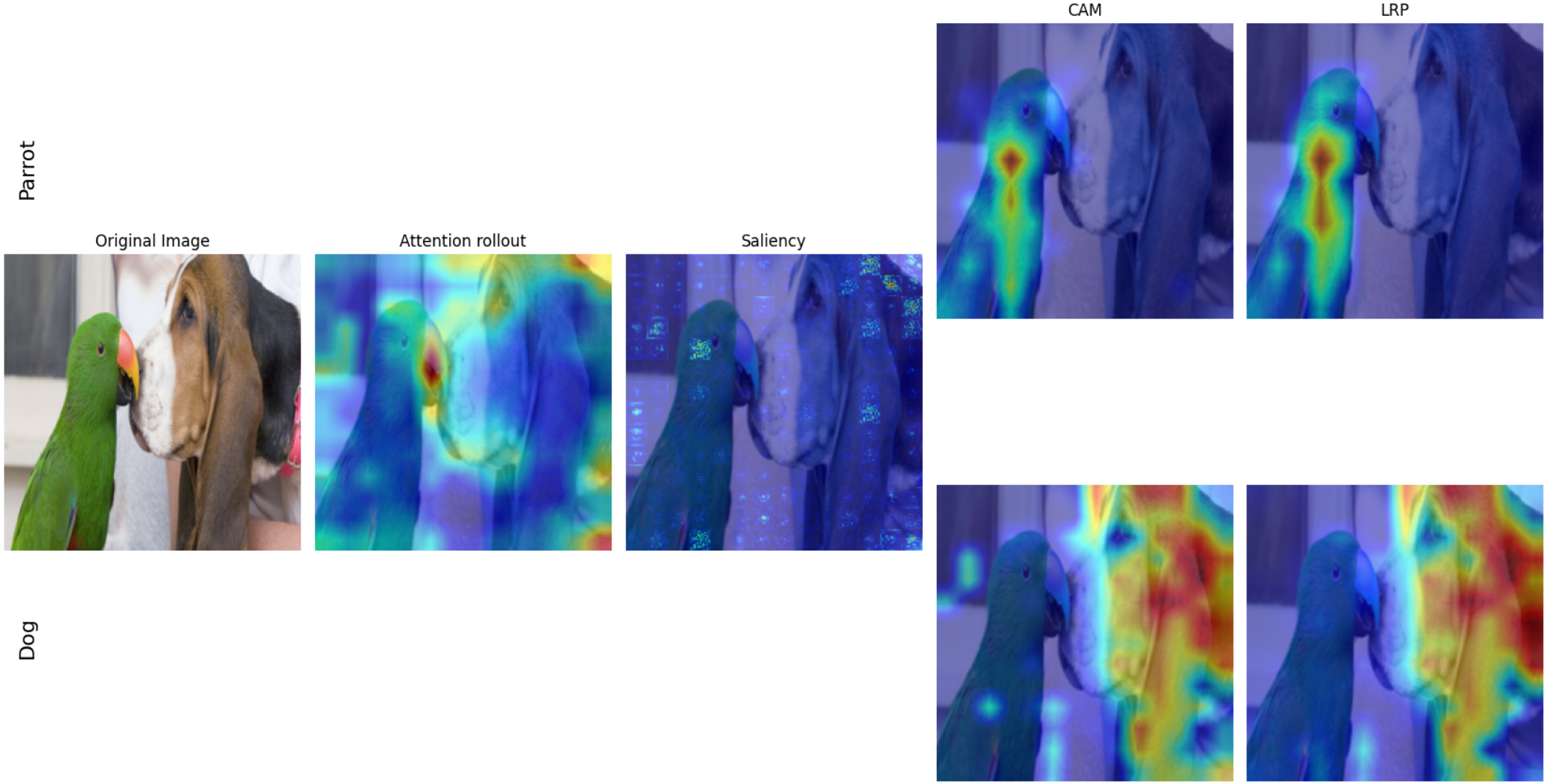}
\caption{Heatmaps of existing visualization techniques}
\label{fig:plot2}
\end{figure}

The second step of this research was to test the hypothesis and see if mixing the different kinds of methods together offers better results. Firstly, LRP was integrated with attention rollout to visualize how attention mechanisms unfold across different layers of the model. This combined method computes relevance scores for each layer's attention head, capturing both the detailed contributions and the evolving focus of the model during decision-making. The mixing of LRP with saliency maps was also explored. By merging these methods, we seek to provide a nuanced understanding of how both global context and local pixel attributes contribute to the model's decision-making process. Secondly, CAM was mixed together with attention rollout, leveraging CAM's ability to highlight discriminative image regions alongside the dynamic attention shifts captured by rollout. This integration aims to provide a comprehensive view of where the model focuses its attention and which regions it deems critical for classification. Lastly, we merge CAM with traditional saliency mapping techniques to jointly highlight salient features identified through gradients and semantic information captured by CAM. This approach aims to balance the interpretability of saliency maps with the spatial context provided by CAM, offering insights into how different features influence the model's predictions across various datasets and tasks.

In evaluating various visualization methods, some clear distinctions emerged. In Figure \ref{fig:plot2} we can see the heatmaps of the four existing techniques that were discussed. As standalone approaches, attention rollout and saliency maps tend to perform poorly, often producing results that are not precise and are prone to significant noise. CAM and LRP, on the other hand, provided much better outcomes, with LRP being particularly effective in highlighting the relevant features more precisely. While CAM results were slightly noisier than LRP, they still outperformed attention rollout and saliency maps. When mixing methods using element-wise multiplication:
\[
\text{transformer\_attribution} = \text{method1} \times \text{method2}
\]
interesting observations were made. The results can be seen in Figure \ref{fig:plot3}. The combinations such as CAM-saliency, LRP-saliency, and LRP-rollout did not exhibit significant improvements over the individual CAM method, suggesting that saliency does not add much extra benefit. However, the combination of rollout with CAM showed a reduction in noise, making the results more similar to those of LRP. This suggests that rollout, when combined with another method, can help highlight the most relevant regions more effectively. Using the geometric mean for mixing methods:
\[
\text{transformer\_attribution} = \sqrt{\text{method1} \times \text{method2}}
\]
showed notable results, which are visible in Figure \ref{fig:plot4}. For LRP, combining it with either attention rollout or saliency highlighted the entire object almost perfectly, indicating its potential utility in object segmentation. 


\begin{algorithm}
\caption{Mixing CAM, LRP, Saliency, and Rollout}
\begin{algorithmic}[1]

\State \textbf{Input:} Feature maps from CAM ($\text{CAM}_i$), LRP ($\text{LRP}_i$), Saliency ($\text{Saliency}_i$), and Rollout ($\text{Rollout}_i$)
\State \textbf{Output:} Combined importance map $M$

\Function{Average}{$X_1, X_2, \ldots, X_n$}
    \State \Return $\frac{1}{n} \sum_{i=1}^{n} X_i$
\EndFunction

\Function{GeometricMean}{$X_1, X_2, \ldots, X_n$}
    \State \Return $\left( \prod_{i=1}^{n} X_i \right)^{\frac{1}{n}}$
\EndFunction

\For{$k = 2, 3$}
        \For{$\text{comb} \in \{\text{all } k\text{-way combinations of CAM,}$ \\
    $\text{LRP, Saliency, Rollout}\}$}

        \State $M_{\text{avg}k} = \text{Average}(\text{comb})$
        \State $M_{\text{geo}k} = \text{GeometricMean}(\text{comb})$
    \EndFor
\EndFor

\State \textbf{Return} $M$

\end{algorithmic}
\end{algorithm}

\section{Model Performance}




The tests were performed on (1) a subset of Pascal VOC, where the predict probability for the main class was higher than 85\%. Such measures were needed as our model was pretrained on ImageNet21k, and we had to ensure the model found the relevant parts in the tested images. The mixing of the methods is summarized in the Algorithm 1. Additionally, we have extended the experiments with (2) PH2 dataset\cite{Mendonca2013PH2}. In the latter, the model started from the same pretrained weights, but we finetuned it with PH2 dataset.

\subsection{Pascal VOC}
Among the individual methods, Rollout, starting from the second layer, achieved the highest scores, with a Jaccard Index of 46.51, an F1 Score of 61.18, and a Pixel Accuracy of 74.81, indicating its strong capability in capturing significant features. LRP also performed well, particularly with the highest Pixel Accuracy of 75.52. CAM and Saliency methods lagged behind, with Saliency showing the lowest performance.

Mixing methods using the geometric mean yielded notable improvements. The LRP and Rollout combination achieved the best results, with a Jaccard Index of 48.36, an F1 Score of 62.34, and a Pixel Accuracy of 78.99, demonstrating the benefit of integrating LRP’s detailed pixel relevance with Rollout’s attention dynamics. Other combinations, like LRP with CAM and Rollout with CAM, also showed enhancements but did not surpass the LRP and Rollout mix. Saliency combinations provided moderate improvements.

Three-way method combinations, also using the geometric mean, generally did not outperform the best two-way combination but still offered valuable insights, indicating possible improvements and a stabler performance.

\begin{table*}[!tb]
\centering
\caption{Average Segmentation Metrics for All Methods. Threshold set to 0.85 accuracy minimum. \textbf{Best results have been bolded}.}
\label{tab:all_methods_85}
\begin{tabular}{lrrr}
\toprule

Method &  Jaccard Index (IoU) &  F1 Score &  Pixel Accuracy \\

\midrule
Single Methods\\
\hline
    GradCAM &                12.43 &     19.13 &           65.94 \\
    LRP &                36.41 &     50.19 &           75.52 \\
    Rollout &                46.51 &     61.18 &           74.81 \\
    Saliency &                 9.18 &     15.63 &           66.25 \\
\hline
Two-way Methods\\
\midrule
    LRP + GradCAM &                22.31 &     32.74 &           69.70 \\
    LRP + Rollout &                \textbf{48.36} &     \textbf{62.34} &           \textbf{78.99} \\
    LRP + Saliency &                35.18 &     49.00 &           73.20 \\
    Rollout + GradCAM &                22.30 &     32.26 &           67.91 \\
    Saliency + GradCAM &                16.40 &     24.28 &           66.50 \\
    Saliency + Rollout &                27.60 &     40.00 &           65.37 \\
\hline
Three-way Methods\\
\midrule
    LRP + Rollout + GradCAM &                32.96 &     45.98 &           73.51 \\
    LRP + Saliency + GradCAM &                25.97 &     37.91 &           70.97 \\
    LRP + Saliency + Rollout &                38.93 &     52.76 &           75.04 \\
    Saliency + Rollout + GradCAM &                21.84 &     31.87 &           68.19 \\
\bottomrule
\end{tabular}
\end{table*}

\subsection{PH2 Dataset}

The extension of our analysis to medical data demonstrates that the hierarchy of results remains consistent, with LRP + Rollout emerging as the best performing method. The mixing strategies further reveal that the 2Way approach outperforms both the simpler 1Way approach and the more complex 3Way variant. Among the combination strategies, the geometric mean consistently proves superior to element-wise multiplication, as evidenced in Tables \ref{tab:comparison_of_1way_methods}, \ref{tab:comparison_of_2way_methods_iou_f1_acc} and \ref{tab:comparison_of_3way_methods_iou_f1_acc}.

\begin{figure}[!tb]
    \centering
    \includegraphics[width=1\linewidth]{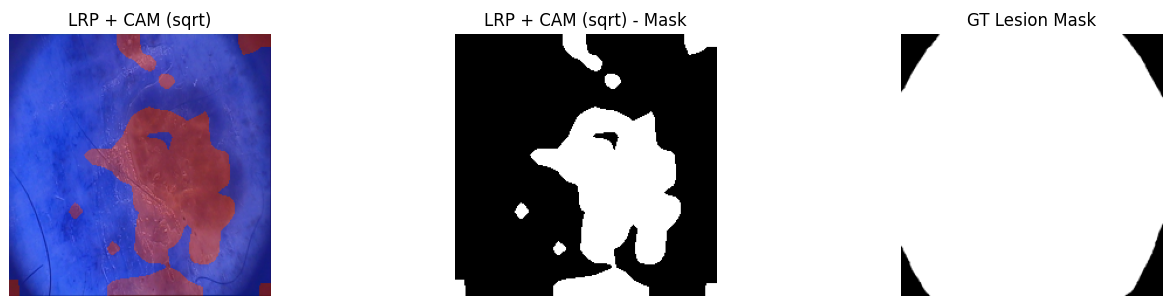}
    \caption{Analysis of the results of the LRP+CAM 2WAY method. While the generated mask appears reasonable at first glance, the ground truth mask in this case seems to be incorrect.}
    
    \label{fig:ph1}
\end{figure}

\begin{figure}[!tb]
    \centering
    \includegraphics[width=1\linewidth]{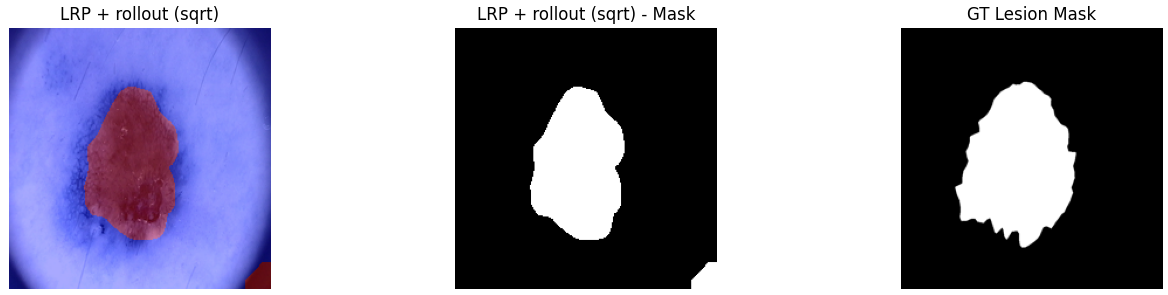}
    \caption{Qualitative example of the segmentation using the best 2Way method.}
    \label{fig:ph2}
\end{figure}

Figures \ref{fig:ph1} present an interesting observation: even when the provided ground truth mask is inaccurate, the explanation methods still effectively highlight regions that intuitively correspond to the area of interest. In contrast, Figure \ref{fig:ph2} shows that compared to an appropriate ground truth mask, our method (specifically LRP + Rollout) produces a segmentation mask that aligns closely with the target region.

\begin{table}[ht]
\centering
\begin{tabular}{lrrr}
\toprule
         Method &  IoU (\%) & F1 Score (\%) & Pixel Accuracy (\%) \\
\midrule
            LRP &                 \textbf{50.47} & \textbf{65.02} & 53.51 \\
       saliency &                 29.89 & 45.29 & 42.73 \\
        rollout &                 31.66 & 52.70 & \textbf{62.91} \\
            CAM &                 23.43 & 19.83 & 16.63 \\
\bottomrule
\end{tabular}
\caption{Comparison of 1WAY Methods (IoU, F1 Score, and Pixel Accuracy).}
\label{tab:comparison_of_1way_methods}
\end{table}

\begin{table*}[ht]
\centering
\begin{tabular}{lrrr}
\toprule
                     Method &  IoU (\%) & F1 Score (\%) & Pixel Accuracy (\%) \\
\midrule
      LRP + saliency (sqrt) &                 54.96 &     67.83 &             59.35 \\
  LRP + saliency (multiply) &                 33.46 &     46.97 &             35.05 \\
       LRP + rollout (sqrt) &                 \textbf{62.09} &     \textbf{70.91} &             63.10 \\
   LRP + rollout (multiply) &                 45.44 &     59.58 &             47.99 \\
           LRP + CAM (sqrt) &                 40.43 &     40.86 &             31.82 \\
       LRP + CAM (multiply) &                 23.76 &     29.77 &             20.20 \\
       rollout + CAM (sqrt) &                 38.17 &     34.40 &             31.66 \\
   rollout + CAM (multiply) &                 21.53 &     21.13 &             17.40 \\
    saliency + CAM (sqrt) &                 35.62 &     34.66 &             30.78 \\
 saliency + CAM (multiply) &                16.79 &     19.40 &             14.26 \\
       saliency + rollout (sqrt) &         43.88 &     59.99 &             \textbf{63.84} \\
 saliency + rollout (multiply) &            29.79 &     43.55 &             37.74 \\
\bottomrule
\end{tabular}
\caption{Comparison of 2WAY Methods (IoU, F1 Score, and Pixel Accuracy)}
\label{tab:comparison_of_2way_methods_iou_f1_acc}
\end{table*}

\begin{table*}[ht]
\centering
\begin{tabular}{lrrr}
\toprule
                                     Method &  IoU (\%) & F1 Score (\%) & Pixel Accuracy (\%) \\
\midrule
      LRP + saliency + rollout (sqrt) &     38.88 &     \textbf{58.15} &            \textbf{54.16} \\
  LRP + saliency + rollout (multiply) &     16.93 &     35.86 &             27.13 \\
          LRP + saliency + CAM (sqrt) &     34.96 &     51.53 &             43.60 \\
      LRP + saliency + CAM (multiply) &     15.64 &     31.37 &             22.37 \\
           LRP + rollout + CAM (sqrt) &     37.48 &     45.77 &             42.04 \\
       LRP + rollout + CAM (multiply) &     17.19 &     30.92 &             24.36 \\
    saliency + rollout + CAM (sqrt) &     \textbf{39.61} &     44.22 &             42.01 \\
 saliency + rollout + CAM (multiply) &     17.77 &     25.81 &             20.00 \\
\bottomrule
\end{tabular}
\caption{Comparison of 3WAY Methods (IoU, F1 Score, and Pixel Accuracy)}
\label{tab:comparison_of_3way_methods_iou_f1_acc}
\end{table*}

\begin{figure*}[!tb]
    \centering
    \includegraphics[width=1\linewidth]{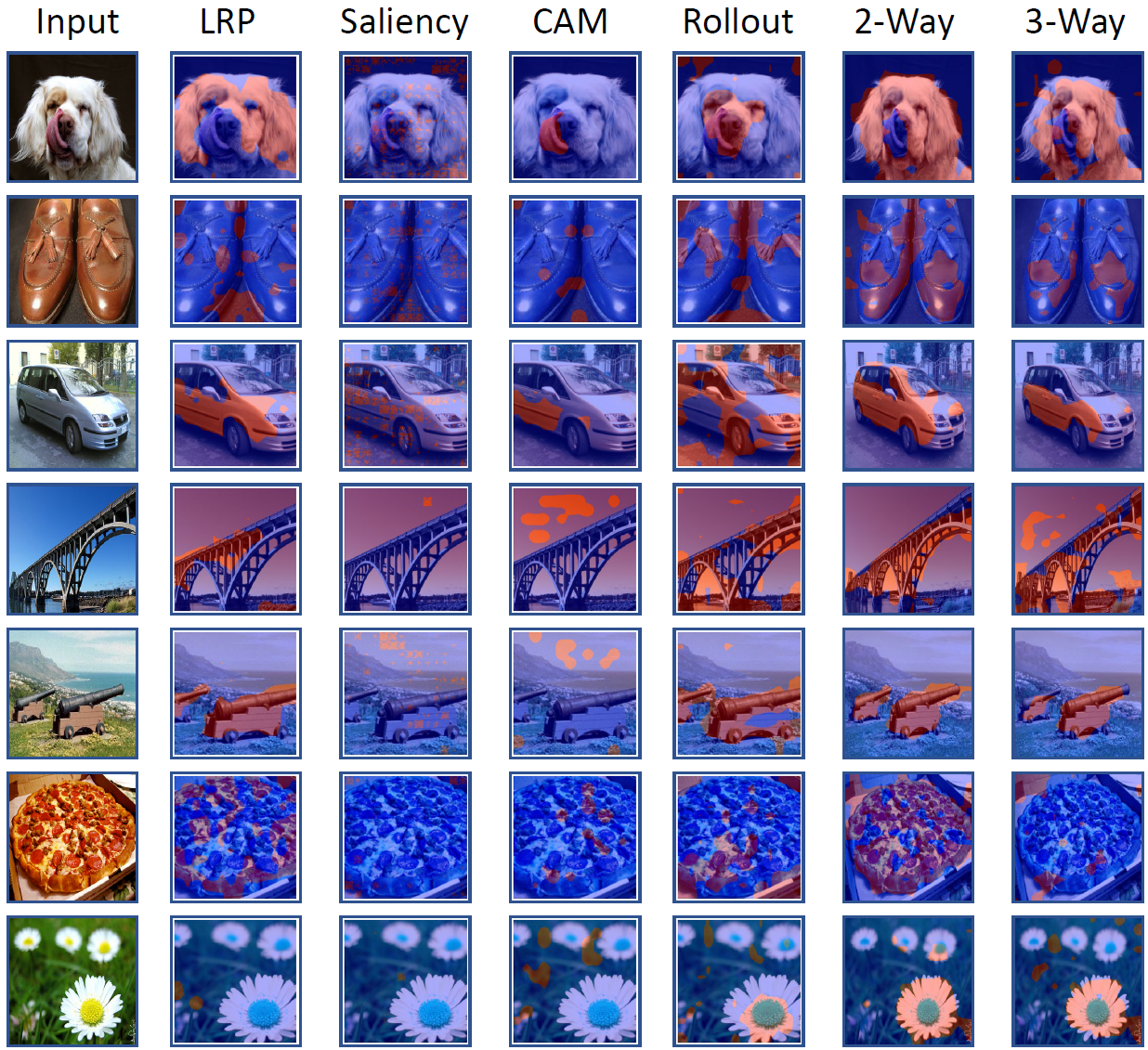}
    \caption{Qualitative results comparison. The baseline methods, from left to right, are LRP, Saliency, Grad-CAM, and Rollout. The columns labeled “2-Way” and “3-Way” display the best results when mixing two and three of the aforementioned methods, respectively.}
    \label{fig:qualitative}
\end{figure*}

\textbf{Model Interpretability}: To accurately assess the effects of the methods mentioned, we selected samples from various ImageNet classes to evaluate the performance of different visualization methods and their optimal combinations. “2-Way” refers to the best-performing combination of two methods, while “3-Way” represents the best combination of three methods. The results are presented in Figure \ref{fig:qualitative}.

\textbf{Explainability gain}: We introduce a novel method to prove the gain in explainability of our model based on the application of the \textbf{Pigeonhole principle}, \cite{pigeon}. This principle states that if you put $n$ pigeons into $k$ holes, then one of the holes contains $\lceil n/k \rceil$ pigeons. To apply it for the geometric
mean of the transformer attributions, we need to reformulate the problem in a linear algebra setting.

Let's consider the explainability feature attribute maps as matrices \( A = [a_{ij}] \) and \( B = [b_{ij}] \) of the same dimensions \( n \times n \). We want to determine if there exists a pair of elements from \( A \) and \( B \) such that their geometric mean is the same as the geometric mean of another distinct pair of elements from the matrices.

\textit{Proof}: Consider the number of possible distinct geometric means that can be formed from pairs of elements from \( A \) and \( B \). For each pair \( (a_{ij}, b_{kl}) \), there is one geometric mean. Since there are \( n^2 \) elements in each matrix, there are \( n^2 \times n^2 = n^4 \) possible pairs of elements. Let \( V \) be the number of distinct possible values that the geometric means can take. \( V \) depends on the specific values of \( a_{ij} \) and \( b_{kl} \), but typically, \( V \) might be significantly smaller than \( n^4 \) because many different pairs can produce the same geometric mean.

The "pigeons" are the \( n^4 \) different geometric means obtained from the pairs \( (a_{ij}, b_{kl}) \). The "pigeonholes" are the distinct possible values these geometric means can take.

If the number of possible pairs \( n^4 \) exceeds the number of distinct geometric mean values \( V \), then by the Pigeonhole Principle, at least one geometric mean value must correspond to more than one pair of elements. This implies that there exist distinct pairs \( (i, j) \) and \( (k, l) \) such that:
        \[
        \sqrt{a_{ij} \cdot b_{kl}} = \sqrt{a_{mn} \cdot b_{pq}}
        \]
        where \( (i, j) \neq (m, n) \) or \( (k, l) \neq (p, q) \).

Under the context-based hypothesis that all features attribution models are visually explaining the same object, the maximum values are concentrated in overlapping regions that are aligned in their barycenter. By considering the geometric mean as a threshold for our post-hoc Pigeonhole-explainabilty measure, it results that the mixed feature attribution map is smoother and has more explainable attributes (\textit{at least one geometric mean value must correspond to more than one pair of elements in the same region}), hence it offers a more precise visualisation than each individual method. 

\section{Conclusions}
\label{sec:conclusion}

This research successfully mixed different explainability techniques for ViTs, uncovering promising methods for both image segmentation and feature visualization. By combining LRP with either saliency maps or attention rollout using the geometric mean, we identified an effective approach for highlighting entire objects, demonstrating potential utility in object segmentation. Furthermore, mixing CAM with attention rollout significantly improved CAM's performance, bringing it on the same level with LRP. Additionally, our observations indicated that attention rollout, when used in combination with other methods, can intensify the highlighted features, providing a more detailed understanding of the model's focus areas.


Mixing multiple explainability visualization techniques helps create a more comprehensive understanding of the input image's features, aiding in the establishment of a hierarchy of the most relevant regions. A formal proof of explainabilty gain is provided and it is based on the application of the Pigeonhole principle. Overall, leveraging multiple methods together can enhance the visualization process, offering deeper insights and more precise interpretations of the model's decision-making process.





\bibliographystyle{unsrtnat}
\bibliography{references}  






\end{document}